\documentclass[review,12pt,3p]{elsarticle}

\usepackage{geometry}
\usepackage{placeins}
\usepackage{xspace}
\usepackage{caption}
\usepackage{subcaption}
\usepackage{graphicx}
\usepackage{url}
\usepackage[hidelinks]{hyperref}
\usepackage{booktabs}
\usepackage{threeparttable}
\usepackage{multirow}
\usepackage{float}
\usepackage{amsmath}

\usepackage{cleveref}
\usepackage{longtable}
\usepackage{natbib}

\usepackage{graphicx}
\usepackage{amssymb}

\usepackage{lineno}
\usepackage[dvipsnames]{xcolor}
\usepackage{ulem}

\definecolor{nyupurple}{RGB}{150, 35, 140}

\journal{arXiv}

\begin{document}

\begin{frontmatter}


\title{Examining spatial heterogeneity of ridesourcing demand determinants with explainable machine learning}



\author[label1]{Xiaojian Zhang}
\address[label1]{Department of Civil and Coastal Engineering, University of Florida}
\ead{xiaojianzhang@ufl.edu}

\author[label1]{Xiang Yan}
\ead{xiangyan@ufl.edu}

\author[label2]{Zhengze Zhou}
\address[label2]{Department of Statistics and Data Science, Cornell University}
\ead{zz433@cornell.edu}

\author[label1]{Yiming Xu}
\ead{yiming.xu@ufl.edu}

\author[label1]{Xilei Zhao\corref{cor1}}
\ead{xilei.zhao@essie.ufl.edu}

\cortext[cor1]{Corresponding author. Postal address: 1949 Stadium Rd, Gainesville, FL 32611, USA.}

\begin{abstract}
The growing significance of ridesourcing services in recent years suggests a need to examine the key determinants of ridesourcing demand. However, little is known regarding the nonlinear effects and spatial heterogeneity of ridesourcing demand determinants. This study applies an explainable-machine-learning-based analytical framework to identify the key factors that shape ridesourcing demand and to explore their nonlinear associations across various spatial contexts (airport, downtown, and neighborhood). We use the ridesourcing-trip data in Chicago for empirical analysis. The results reveal that the importance of built environment varies across spatial contexts, and it collectively contributes the largest importance in predicting ridesourcing demand for airport trips. Additionally, the nonlinear effects of built environment on ridesourcing demand show strong spatial variations. Ridesourcing demand is usually most responsive to the built environment changes for downtown trips, followed by neighborhood trips and airport trips. These findings offer transportation professionals nuanced insights for managing ridesourcing services. 




\end{abstract}


\end{frontmatter}



\section{Introduction}
\label{S:1}
The rapid expansion of ridesourcing services (e.g., Uber and Lyft) in recent years has significantly changed how people travel \citep{ghaffar2020modeling}. As ridesourcing services become an increasingly important component of the transportation ecosystem, transportation policymakers need to learn how to effectively manage this new mobility option. To develop effective policy strategies that promote the positive impacts of ridesourcing while mitigating its negative effects, transportation professionals need better knowledge of what factors and how do they affect ridesourcing demand.

A number of recent studies have explored which factors (e.g., built environment variables) are associated with ridesourcing demand \citep{ghaffar2020modeling, marquet2020spatial, lavieri2018model}. These studies have commonly assumed these associations to be linear or log-linear. However, this assumption has been challenged by some studies employing machine-learning methods, which show that the associations between ridesourcing demand and its key determinants are often nonlinear \citep{yan2020using, xu2021identifying, tu2021exploring}. In addition to being supported by empirical results, nonlinear relationships such as threshold effects can make better intuitive or theoretical sense in many cases. Consider the example of how density shapes travel demand. While the common finding is that travel demand rises with density increases \citep{ewing2010travel}, the impact is expected to have an effective range: travel demand is only responsive to density increases if density is beyond a certain level (i.e., the effect does not start at a density value of zero), and this positive influence is likely disappear when density reaches a upper threshold and travel demand becomes ``saturated" \citep{cheng2021examining}. Therefore, identifying and understanding these nonlinear relationships (e.g., threshold effects and effective range) has important implications for transportation planning and decision-making \citep{ding2021non, shao2020threshold}. 


A further complicating factor in understanding and modeling ridesourcing demand is the issue of spatial heterogeneity; that is, what factors and how these factors influence ridesourcing demand can differ across spatial contexts. Some previous studies that model the spatial variations of travel demand have demonstrated the importance of specifying spatially varying coefficients \citep{qian2015spatial,yu2019exploring}. Several studies that apply machine learning (ML) approaches to model nonlinear relationships further suggest that the nonlinear effects of built environment on travel behavior (e.g., transit ridership \citep{chen2021nonlinear} and commuting mode choice \citep{ding2021non}) vary across locations. A recent study of ridesourcing trips in Chicago has further shown that unique compositions of individual and environmental characteristics lead to evident variations in the patterns of ridesourcing trips in downtown, airport and neighborhood \citep{zhang2022machine}. Therefore, to accurately predict ridesourcing trips, it not only requires a clear understanding of which factors affect ridesourcing demand and their nonlinear associations, but also demands explicit efforts in accounting for spatial heterogeneity in the modeling process. However, to our knowledge, no existing study on ridesourcing demand modeling has simultaneously modeled nonlinear effects and spatial heterogeneity regarding the key determinants (e.g., built environment factors) of ridesourcing demand. 

This study fills the research gap by leveraging the publicly available ridesourcing-trip data in Chicago. We take the following steps. First, we adopt a machine learning method (i.e., XGBoost) to model the ridesourcing demand at the Census Tract level. This model can effectively capture the nonlinear relationships between the explanatory variables and the outcome. Second, we identify three types of ridesourcing trips, i.e., downtown, airport and neighborhood trips. The built environment or sociodemographic characteristics are significantly different in downtown, airport and neighborhood areas \citep{zhang2022machine}, providing an opportunity to investigate how the impacts of key factors associated with ridesourcing demand vary across different spatial contexts. Third, We apply an emerging ML explanation tool, SHapley Addictive exPlanations (SHAP) \citep{lundberg2017unified,lundberg2020local}, to identify which factors contribute more to predicting ridesourcing demand patterns and how these results vary across spatial contexts (i.e., downtown, airport and neighborhood). This approach enhances our understanding of which factors should be included when modeling ridesourcing demand at various spatial contexts. Finally, we apply \textit{Conditional Partial Dependence Plot} to interpret the associations between ridesourcing demand and its determinants, focusing on identifying nonlinear effects and spatial heterogeneity in these effects. Our study results challenge assumptions of linearity and spatial homogeneity and instead suggest that several key factors have nonlinear associations with ridesourcing demand and these associations vary across spatial contexts. These results can inform transportation policymakers to make targeted policies and strategies to facilitate the management of ridesourcing services.


The remaining paper is structured as follows: Section \ref{LR} reviews the related studies. The following sections introduce the data and the methodological framework. Section \ref{results} presents the results. Section \ref{discussion and conclusion} concludes this study with a summary of key findings and a discussion of limitations and future research directions. 

\section{Literature Review}
\label{LR}

\subsection{Determinants of ridesourcing demand}

The ridesourcing services have received great attention in the transportation community in recent years. To date, a large body of efforts has been devoted to identifying the determinants of the demand for ridesourcing services. In its early stage, due to the limited availability of ridesouring-trip data, most of the studies are conducted by taking interviews and surveying. For example, \citet{rayle2016just} analyzed 380 intercept surveys collected in San Francisco and found that ridesourcing are attractive to generally younger and well-educated travelers. \citet{sarriera2017share} found that race, gender and income level are important factors that could influence rider's willingness to use ridesourcing. Since the surveys are usually conducted at the disaggregated (i.e., individual) level, we can directly derive insights on the underlying behavioral mechanism. However, conducting surveys or interviews takes great monetary resources and is usually time-consuming. In recent years, more and more researchers have started to explore the determinants of ridesourcing demand through publicly released real-world ridesourcing-trip data. \citet{xu2021identifying} studied the relationships between ridesplitting adoption rate and its determinants using Chicago ridesourcing-trip data. Empirical results suggested that ethnic composition, education attainment, income level, transit accessibility and Neighborhood density are critical factors in shaping ridesplitting adoption rate. \citet{ghaffar2020modeling} used the same data source and found that parking spot availability, land-use diversity and criminal rates play important roles in shaping ridesourcing trip generations. \citet{yan2020using} reached the similar conclusions in the context of Chicago; but their results also indicated that travel impedance attributes are most important in predicting ridesourcing demand. Most of the other studies collectively suggested that socioeconomic and demographic, built environment and transit supply variables could significant shape ridesourcing demand. Overall, these studies provide valuable instructions on which factors should we select in modeling the ridesourcing demand.

Although these variables collectively display a significant role in shaping ridesourcing demand, the conclusions of how and to what extent do they act remain debated among literature. Empirical results from recent research reported that built environment variables could have contrary effects on ridesourcing demand. For example, \citet{lavieri2018model} found that the retail density is positively associated with trip-generation rate of ridesourcing services. However, \citet{marquet2020spatial} concluded that their relationship is positive. Several studies revealed that population density and employment density has a positive relationship with ridesourcing demand \citep{ghaffar2020modeling, lavieri2018model}. However, \citet{yu2019exploring} presented that the population density can both positively and negatively impact ridesourcing demand depending on locations; and the effect of employment density could be negative for areas with high income level \citep{tu2018spatial}. There are also conflicting findings regarding household car ownership. \citet{marquet2020spatial} observed that the more presence of the no-car households, the fewer ridesourcing-trip generations. By contrast, \citet{lavieri2018model} identified a positive correlation between household car (automobile) ownership with ridesourcing demand. The reverse relationships were also found among transportation supply variables \citep{marquet2020spatial, yan2020using, lavieri2018model}. These contradicting findings could be possibly attributed to the existence of \textit{spatial heterogeneity}, which refers to the spatial variations in the relationships between the explanatory variables and the outcome \citep{fotheringham2003geographically}. Spatial heterogeneity results from the varying patterns of urban forms, which is composed of location-specific individual and environmental characteristics \citep{cheng2021examining}. In recent years, transportation planners have gradually started to recognize the importance of considering spatial heterogeneity in travel behavior research, due to the following reasons. First, models that account for spatial heterogeneity usually enjoy better predictive accuracy \citep{zhang2022machine, chen2021nonlinear}. Second, generating the local information (e.g., how one built environment element is associated with travel demand across spatial units) can help transportation planners promote the understanding of location-specific effects and then develop targeted policies to effectively regulate the role of each travel mode in urban transportation ecosystem \citep{cheng2021examining, ding2018joint}. Therefore, it is of great necessity to consider the influence of spatial heterogeneity and explore the variables' heterogeneous effects on ridesourcing demand.

\subsection{Approaches to identifying spatial heterogeneity}

In order to cope with the spatial variations, several studies applied geographically weighted regression and its extensions (e.g., geographically weighted Poisson regression and geographically and temporally weighted regression) into travel demand modeling \citep[e.g.,][]{yu2019exploring, ma2018geographically, tu2018spatial, qian2015spatial, chen2019discovering}. For example, \citet{tu2018spatial} used GWR to evaluate the relationship between public ridership and socio-demographics, land-use and transportation-supply variables. Results showed that the variable parameters could vary across different contexts. \citet{chen2019discovering} compared the performance of GWR and OLS in metro ridership modeling and found that GWR yielded better fitting results in all scenarios. Theoretically, GWR can efficiently deal with spatial nonstationarity and allows local parameter estimations \citep{brunsdon1996geographically}, hence possessing strong strength in addressing spatial dependencies of the travel demand. However, GWR cannot capture the nonlinearity because it has a predetermined model structure; and this drawback further limits its predictive accuracy \citep{tu2021exploring}. 


Machine learning has been broadly applied to explore the complex relationships behind the travel behavior, due to its superiority in predictive accuracy and dealing with nonlinearity \citep[e.g.,][]{xu2021identifying, yan2020using, ding2019does}. However, there is a growing recognition that machine learning models can hardly consider spatial heterogeneity \citep{chen2021nonlinear, ding2021non}. One way to tackle this problem is to build location-specific submodels which take spatial heterogeneity into account. For example, \citet{chen2021nonlinear} integrated the idea of GWR into the random forest model. The hybrid model incorporates the geographical weights of observations into building each local random forest. Therefore, it can efficiently capture the spatial heterogeneity and address the nonlinear effects over space. \citet{zhang2022machine} proposed to use clustering techniques to capture the spatial heterogeneity and then build cluster-specific model separately for each spatial group. The modeling results demonstrated that this approach effectively addresses the influence of spatial heterogeneity and thus significantly improves the ridesourcing demand predictions. These two methods are theoretically similar since they both try to address the spatial heterogeneity through further polishing machine learning models. However, they have complex model structures (e.g., integration of several models; multiple hyperparameters to tune); and sometimes even need to be facilitated by human intelligence \citep{zhang2022machine}.  

Another popular recognition is that machine learning could automatically capture the heterogeneity behind the data \citep{lheritier2019airline}. Against this backdrop, developing explanation tools that could directly extract the heterogeneity from the black-box machine learning models is promising. In this way, we can not only maintain the prediction accuracy and the model simplicity, but also uncover the spatial variations in the nonlinear associations. Previous studies frequently use variable importance and partial dependence plot to interpret machine learning models \citep{molnar2020interpretable}. Researchers usually use these two methods to decide the predictive importance of the variables and their correlations with the outcome \citep{xu2021identifying, yan2020using}. However, these two methods are not able to directly identify spatial heterogeneity. There is still limited analytical framework available for local explanation purposes, especially for travel behavior research. This study fills the knowledge gap by (1) applying SHAP, an emerging interpretation technique designed for offering local explanations \citep{lundberg2017unified, lundberg2020local}, to generate specific variable importance for each spatial context; (2) adopting conditional partial dependence plot (CPDP) to reveal the (nonlinear) relationships across spatial contexts \citep{zhao2019modeling, molnar2020model}.

\section{Data}
\label{data}
We collected the publicly available ridesourcing-trip data from the City of Chicago Open Data Portal. The data used here are from November 1, 2018 to March 31, 2019, containing 45,338,599 trips. The key attributes of the trip data used here include trip fare, trip distance, duration and pick-up/drop-off locations. Due to privacy concerns, the City of Chicago has rounded the fare and time to the nearest \$2.50 and 15 minutes, respectively; and instead of providing the exact latitude/longitude, the city has aggregated the pick-up/drop-off locations at the census tract level. We followed \citet{yan2020using, xu2021identifying, zhang2022machine} when preparing the data for analysis and modeling. Specifically, we first aggregated the data from a disaggregated level (i.e., individual trip) to an aggregated level (i.e., OD pair). The data provider has randomly removed the trip origins or destinations for some observations to protect the traveler's privacy. Therefore, a missing-value imputation approach is adopted to infer the trip origins and destinations \footnote{The detailed description can be found in \citet{xu2021identifying}}. We also removed the outliers. For example, we excluded OD pairs with 50 trips or less over the study period to minimize the potential impact of randomness. This study did not consider trips that took place on weekends and federal holidays. After removing the outliers, we calculated the average number of trips per day and the following six travel impedance variables: median of trip fare, distance and duration, standard deviation of trip fare, distance and duration. 

We further supplemented the ridesourcing-trip data with socioeconomic and demographic data from American Community Survey (ACS) 2013–2017, several built environment variables developed from the 2015 Longitudinal Employer-Household Dynamics (LEHD) data, and a few transit supply variables developed from the General Transit Feed Specification (GTFS) data. The data preparation steps create 67,498 OD pairs for analysis, including 752 census tracts as trip destinations and 755 census tracts as trip origins. The final dataset includes 30 explanatory variables and one outcome variable (i.e., the average number of trips per day). Travel impedance variables are at the OD-pair level while other variables such as built environment are collected at both trip origins and destinations. Table \ref{tab:Variable Description} presents the variable description.

\begin{figure}[!ht]
    \centering
    \includegraphics[width =\textwidth]{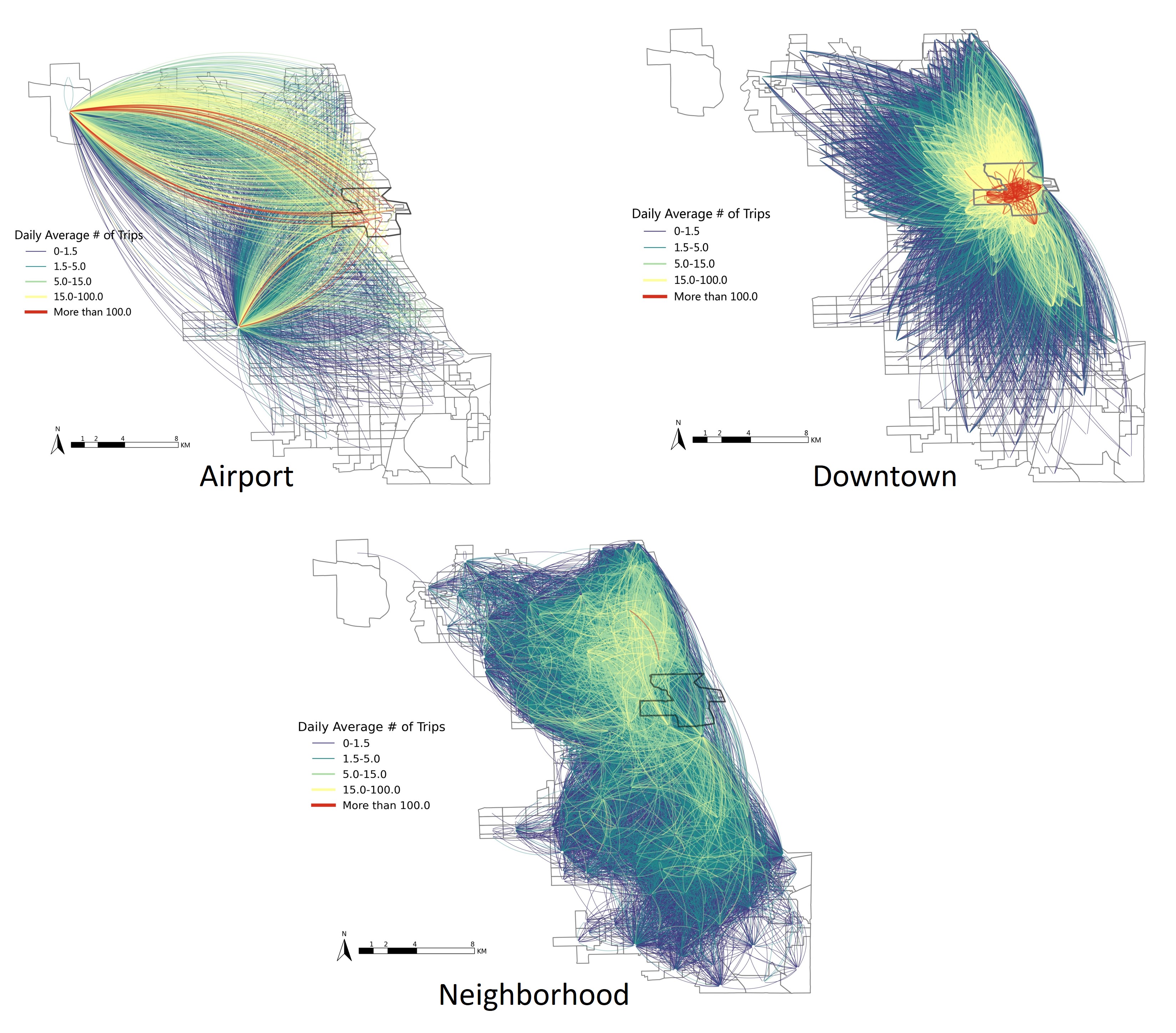}
    \caption{Spatial Distribution of three contexts}
    \label{fig:distribution_sampled_trips}
\end{figure}

Previous studies have found that the ridesourcing-trip demand in Chicago has a strong spatiotemporal disproportion. For example, \citet{zhang2022machine} indicated that ridesourcing travels concentrated at the downtown and north part of Chicago; and the quantity of airport rides was large. Similarly, \citet{ghaffar2020modeling} found that the daily trips in Chicago significantly vary across neighborhoods and over time. These findings further encourage us to explore which factors shape the ridesourcing demand and quantify their heterogeneous effects over space. Therefore, we divided ridesourcing trips into three groups based on the spatial contexts , i.e., \textit{Neighborhood OD Pairs} \textit{Downtown OD Pairs} and \textit{Airport OD Pairs}, from the dataset for analysis. The spatial distribution of the OD pairs is shown in Fig. \ref{fig:distribution_sampled_trips}. We mark the downtown census tracts with more highlighted black lines for their boundaries. We also classify the OD pairs based on the trip volume for each spatial context. Results show that most of OD pairs inside downtown areas and OD pairs connecting downtown and airports have more than 100 ridesourcing-trip demand per day, while OD pairs in neighborhood have relatively lower ridesourcing-trip demand per day.

\begin{table}[!ht]
\centering
\caption{Variable Description}
\label{tab:Variable Description}
\resizebox{\textwidth}{!}{%
\begin{tabular}{ll}
\hline
Variable                                & Description                                    \\ \hline
\textbf{Target variable}                          &                                                \\
Average\_number\_of\_trips\_per\_day                              & The average number of ridesouring trips per day                \\
\textbf{Travel impedance}               &                                                \\
Fare\_median                                     & Median trip cost (\$)                    \\
Fare\_sd                                         & The standard deviation of trip cost (\$) \\
\textbf{Socio-economic and demographic} &                                                \\
Pctsinfam                                         & Percentage of single-family homes              \\
Pcthisp                                           & Percentage of Hispanic population              \\
Pctasian                                          & Percentage of Asian population                 \\
Pctlowinc  & Percentage of low-income households (\$25 k less)                                \\
CrimeDen                                          & Density of violent crime                       \\
Commuters  & Total number of commuters (from origin census tract to destination census tract) \\
\textbf{Built environment}              &                                                \\
\textit{Land use and accessibility}               &                                                \\           
Popden                                            & Population density                             \\
IntersDen                                         & Intersection density                           \\
EmpDen                                            & Employment density                             \\
EmpRetailDen                                      & Retail employment density                      \\
\textit{Transit supply}                 &                                                \\
SerHourBusRoutes                                  & Aggregate service hours for bus routes         \\
PctRailBuf & Percentage of tract within 1/4 mile of a rail stop                               \\
BusStopDen                                        & Number of bus stops per square mile            \\
RailStationDen                                    & Number of rail stops per square mile           \\ \hline
\end{tabular}%
}
\end{table}

\section{Approach}
\label{research design}
This study employs XGBoost and two explanation tools (i.e., SHAP and partial dependence plot) to model the ridesourcing demand and explore the importance of each built environment in forecasting ridesourcing demand and their relationships across three trip types. We provide a detailed description of these approaches below.

\subsection{Modeling Ridesourcing Demand}

We adopt XGBoost to model ridesourcing demand at the Census Tract level in Chicago. Introduced by \citet{chen2016xgboost}, XGBoost can cope with high-dimension input variables in noisy environments and produces high-level prediction accuracy, thus has been widely used in travel behavior modeling \citep{zhang2022machine, liu2021non}. It builds up the boosted decision trees successively with high efficiency and it can be operated in parallel, thus enjoying superior computational speed. In this study, suppose we use a decision tree as the base learner for XGBoost and we have training data $\mathcal{D} = \{Z_{i} = (X_{i}, Y_{i}), i = 1,2, \ldots, n\}$, with $X$ representing the input features and $Y$ representing the ridesourcing demand. When deciding the input features for the model, we excluded candidate variables with a Variance Inflation Factor (VIF) $>$ 10 to mitigate the issue of multicollinearity \citep{xu2021identifying,yan2020using}. Suppose the trees are built on $m$ dimensions of features and $K$ additive functions, then the XGBoost model can be formulated as:

\begin{equation}
\widehat{Y}_{i} = \hat{f}(X_{i})
=\sum_{k=1}^{K} f_{k}\left(X_{i}\right), \quad f_{k} \in \mathcal{F}
\end{equation}

\noindent where $\mathcal{F}=\left\{f(X)=w_{q(X)}\right\}\left(q: \mathbb{R}^{m} \rightarrow T, w \in \mathbb{R}^{T}\right)$ represents the space of all regression trees; $q$ refers to the structure of each tree and $T$ is the number of leaves in each tree. $f_{k}$ represents an individual tree configured with structure $q$ and leaf weights $w$. 

In this study, we tuned the XGBoost with \textit{Grid Search} and a 5-fold \textit{Cross Validation}. Specifically, we examined the number of trees by setting the values from 100 to 200 at an interval of 10; we tested the learning rate by setting the value ranging from 0.01 to 0.05 with a step of 0.01. Consequently, the XGBoost was built with 200 trees with a learning rate of 0.05. The depth of the tree is set as 5 to prevent the modeling from overfitting.

\subsection{Interpreting the XGBoost model}
To identify the determinants of ridesourcing demand and their associations over space, we further interpret the XGBoost model with two ML explanation tools: SHapley Addictive exPlanations (SHAP) and Conditional Partial Dependence Plot (CPDP). Specifically, we use SHAP to uncover the key determinants of ridesourcing demand and use CPDP to show their associations across three spatial contexts.

SHAP is an emerging ML technique in evaluating interpretations in local analytical context. Based on game theory, SHAP can fairly capture the contribution of each predictor and reflect to what extent can each predictor impacts the outcome variable \citep{lundberg2017unified, lundberg2020local}. Besides, SHAP can be applied in various local contexts, including both individual-specific interpretation and subgroup-specific interpretation \citep{xiao2021nonlinear, molnar2020interpretable}. By aggregating the individual interpretations, SHAP can also reflect a global pattern of model behavior \citep{xiao2021nonlinear, parsa2020toward}. Furthermore, SHAP can be adopted to directly examine the variable interactions \citep{lundberg2017unified, lundberg2020local}, which may boost the discovery of heterogeneous effects among different contexts. In this study, we can use SHAP to generate the context-specific variable importance. This approach remedies the limitation of several ML explanation tools for only providing a global measure of variable importance. In SHAP, the contribution of each variable is represented by its Shapley Value \citep{shapley201617}. For every individual prediction, the Shapley Value is obtained by averaging the the contribution of a variable through all coalitions (i.e., feature value combinations). Specifically, the Shapley Value $\phi$ of a feature $j$ is determined as:

\begin{linenomath}
\begin{equation}
    \phi_{j}=\sum_{S \subseteq F \backslash\{j\}} \frac{|S| !(|F|-|S|-1) !}{|F| !}\left[f_{S \cup\{j\}}\left(x_{S \cup\{j\}}\right)-f_{S}\left(x_{S}\right)\right],
    \label{Eq: shapley}
\end{equation}
\end{linenomath}
\noindent where $S$ is the feature subset, $F$ is the set of all feature, $f_{S \cup\{j\}}$ and $f_{S}\left(x_{S}\right)$ represent the model trained with and without feature $j$, respectively. After the Shapley Value for all variable are obtained, we can calculate the relative absolute impact of each variable on the outcome:

\begin{linenomath}
\begin{equation}
    P_{j}=\frac{1}{n} \sum_{i=1}^{n}\left|\phi_{j}^{(i)}\right|,
    \label{Eq: shap impact}
\end{equation}
\end{linenomath}
\noindent where $P_{j}$ is the relative absolute impact for variable $j$ and $\phi_{j}^{(i)}$ is the Shapley Value for variable $j$ in individual prediction $i$. Sequentially, the relative variable importance can be formulated by standardizing all variables' relative absolute impact:

\begin{linenomath}
\begin{equation}
    I_{j}=\frac{1}{N} \sum_{j=1}^{N}\left|P_{j}\right|,
    \label{Eq: shap importance}
\end{equation}
\end{linenomath}
\noindent where $I_{j}$ denotes the relative variable importance for variable $j$, $N$ is the total number of features used for analysis. Accordingly, this measure can be directly applied in a context (e.g., for all individual prediction $i$ belonging to subgroup $g$) to obtain the context-specific relative variable importance.

The variable importance refers to the relative contribution of a feature to the predictive power of the modeling results, which is comparable to the magnitude of the beta coefficient in conventional statistical models. Therefore, it cannot indicate the direction (i.e., sign) of the variable's relative contribution.

Prior research usually applied Partial Dependence Plot (PDP) to reveal the direction of associations between the predictors and the target variable \citep[e.g.,][]{xu2021identifying, chen2021nonlinear}. Introduced by \citet{friedman2001greedy}, PDP can display the marginal effect that the studied variable has over the predicted outcome \citep{molnar2020interpretable}. Specifically, suppose the set $S$ contains the features of interest and set $C$ is the complement of $S$ and let $x$ denote the predictors, the PDP works by marginalizing the predicted outcome over $x_c$. The partial dependence of the model $f$ on $x_S$ is defined as:

\begin{linenomath}
\begin{equation}
    \hat{f}_{x_{S}}\left(x_{S}\right)=E_{x_{C}}\left[\hat{f}\left(x_{S}, X_{C} \right)\right]
    \label{Eq: pdp1}
\end{equation}
\end{linenomath}

In this study, $\hat{f}_{x_{S}}$ is estimated by taking the average of the data samples, as shown in Eq. \ref{Eq: pdp2} where $n$ is the number of total instances:

\begin{linenomath}
\begin{equation}
    \hat{f}_{x_{S}}\left(x_{S}\right)=\frac{1}{n} \sum_{i=1}^{n} \hat{f}\left(x_{S}, x_{C}^{(i)} \right)
    \label{Eq: pdp2}
\end{equation}
\end{linenomath}

With PDP, we are able to investigate the average marginal effect of a given value of feature in $S$ on the ridesourcing demand. However, one basic assumption of PDP is that the studied feature is not correlated with the others. High interactions among features may hide the heterogeneous effects, resulting in a misleading interpretation \citep{molnar2020interpretable, molnar2020model, zhao2019modeling}. To address this issue, \citet{goldstein2015peeking} proposed the Individual Conditional Expectation (ICE) plot. Instead of generating the average marginal effect, the ICE plot can reveal the partial dependence of a feature for each instance separately and the average of all ICE curves is equivalent to the PDP \citep{molnar2020interpretable}. Focusing on the individual-specific interpretation, ICE plot helps uncover the heterogeneity potentially existed in the data. Specifically, the ICE curve for a particular instance $i$ is:

\begin{linenomath}
\begin{equation}
    \hat{f}_{S}^{i}\left({x}_{S}\right) = \hat{f}\left(x_{S}^{i}, x_{C}^{i}\right)
    \label{Eq: ice1}
\end{equation}
\end{linenomath}

Even though the PDP and ICE plots enable us to extract valuable insights from black-box machine-learning models, they can only be applied in a global analytical context (i.e., across all data samples). However, we aim to explore how the heterogeneous effects of the key determinants can vary over spatial contexts. Hence, the approaches dedicated for a specific context (e.g., subgroup or segment) of data samples are required. 

To complement the application restrictions of global methods, several recent works developed subgroup-specific interpreting techniques for context-specific analyzing purpose \citep{zhao2019modeling, molnar2020interpretable}. Among these, two innovative approaches are Conditional Individual Partial Dependence Plot (CIPDP) and Conditional Partial Dependence Plot (CPDP). Specifically, the principal idea of CIPDP and CPDP is interpreting model results under certain \textit{Conditions}, e.g., a subgroup of data samples or an individual prediction. By accounting for the subgroup-specific variations, these plots further allow researchers to explore the heterogeneous effects of the predictors over different contexts \citep{zhao2019modeling}. Considering we can divide the original dataset into $G$ subgroups, where $G=\{g_1, g_2,\ldots, g_n\}$, the CPDP conditioned on a subgroup $g$ is specified as: 

\begin{linenomath}
\begin{equation}
    \hat{f}_{x_{S}}\left(x_{S}\right)_{\mid x \in g}=E_{x_{C}}\left[\hat{f}\left(x_{S}, x_{C},x \in g \right)\right] = \frac{1}{n_g} \sum_{i=1}^{n_g} \hat{f}\left(x_{S}, x_{C}^{(i)},x\in g\right),
    \label{Eq: cpdp}
\end{equation}
\end{linenomath}
\noindent where $n_g$ is the number of instances included in subgroup $g$.

CIPDP is an extension to ICE plots; and similarly, the average of all CIPDP curves is equivalent to CPDP. The CIPDP for an individual instance $j$ in subgroup $g$ can be formulated as:

\begin{linenomath}
\begin{equation}
    \hat{f}_{S}^{j}\left({x}_{S}\right)_{\mid x\in g} = \hat{f}\left(x_{S}^{j}, x_{C}^{j}, x\in g\right)
    \label{Eq: cice}
\end{equation}
\end{linenomath}

In this study, we first generate all CIPDPs and then calculate the corresponding CPDP to show the spatially-varying relationships between different determinants and ridesourcing demand.

\section{Results}
\label{results}

Here we first briefly introduce the descriptive statistics of three spatial contexts. Then, we discuss the variables that contribute the most to predicting the three types of ridesourcing trips (airport, downtown, and neighborhood trips). Finally, we examine the CPDP curves to identify possible nonlinear associations between several factors that have high importance values and ridesourcing demand. Our discussion will focus on effective range and threshold effects, and we will compare how these nonlinear effects differ across spatial contexts.

\subsection{Descriptive Statistics}

Table \ref{tab:Descriptive Statistics} (in appendix) presents the descriptive statistics for each variable across three spatial contexts. We observe that for variables examined at both the trip origin and destination, their corresponding characteristics are quite similar. Moreover, airport context has the highest average ridesourcing-trip demand per day, followed by downtown and neighborhood. Regarding the land use and accessibility variables, trip origins and destinations in downtown have extremely large employment density, retail employment density, population density and intersection density. This finding also corresponds to the largest volume of commuters in downtown context. Airport context has the poorest transit supply, indicating trip origins and destinations for airport trips are not well served by public transit. These spatially varying characteristics lead to a distinguishable ridesourcing demand pattern across three contexts.

\subsection{Variable Importance}

\begin{figure}[!ht]
    \centering
    \includegraphics[width=\textwidth]{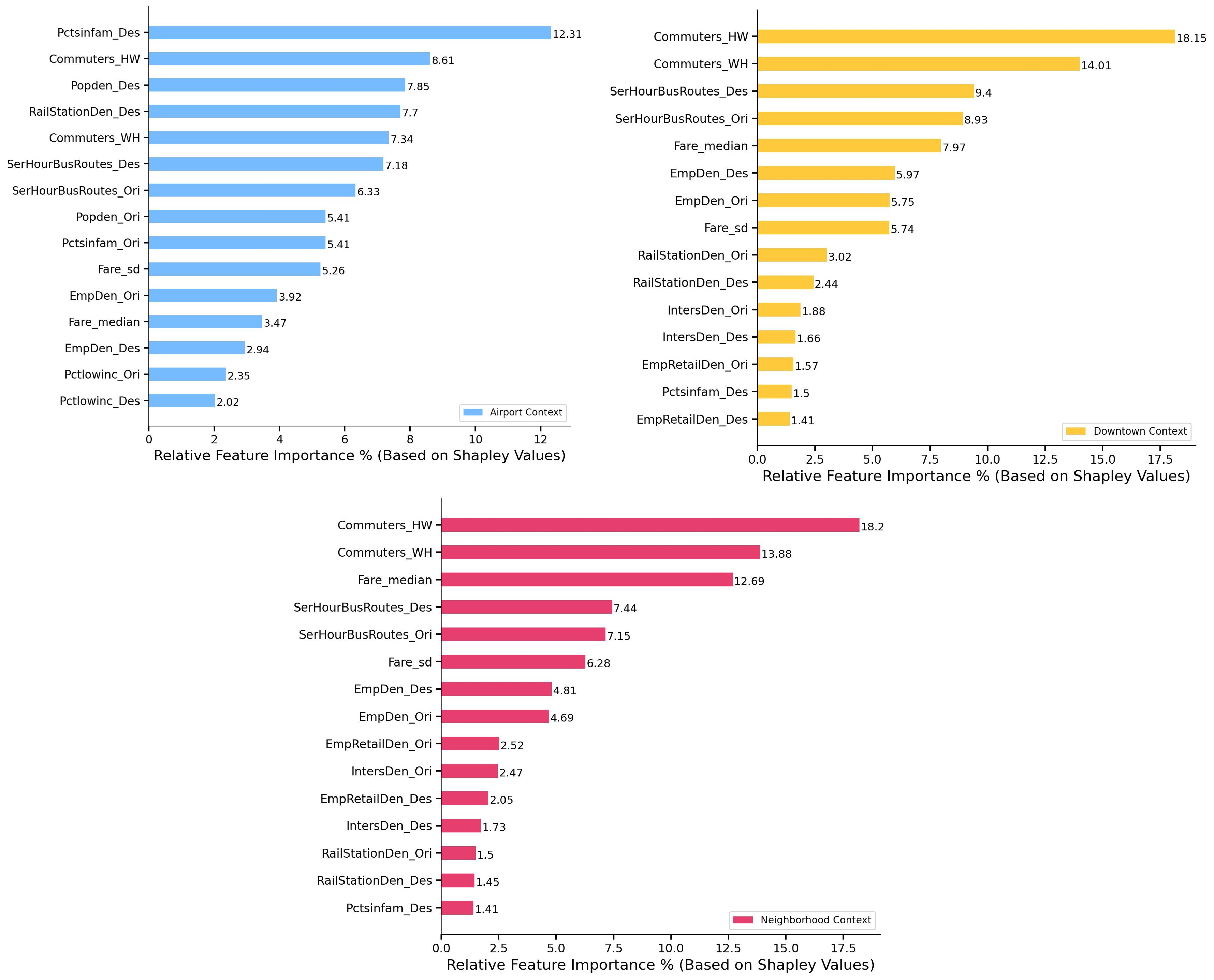}
    \caption{Variable Importance}
    \label{fig:vi}
\end{figure}

For tree-structured models, the global feature importance is usually computed by the mean decrease in node impurity when splitting on this variable \citep{breiman2001random}. However, this measure cannot be applied in a local context. This study calculates the context-specific feature importance by averaging the absolute Shapley Value for each variable. We present the relative variable importance for the top15 variables in three spatial contexts.

\begin{table}[!ht]
\centering
\caption{Relative Importance of Variables by Category}
\label{tab:category_importance}
\resizebox{\textwidth}{!}{%
\begin{tabular}{@{}llllllll@{}}
\toprule
\multirow{2}{*}{Variable Category} & \multirow{2}{*}{Variable Count} & Importance Sum. &  &  & Importance Avg. &  &  \\ \cmidrule(l){3-8} 
                               &    & Neighborhood & Downtown & Airport & Neighborhood & Downtown & Airport \\ \cmidrule(l){2-8} 
Travel Impedance               & 2  & 18.97          & 13.71    & 8.73    & 9.49           & 6.86     & 4.36    \\
Socio-Economic and Demographic & 12 & 40.39          & 39.01    & 40.93   & 3.37            & 3.25     & 3.41    \\
Built Environment              & 16  & 40.64         & 47.28    & 50.35   & 2.54           & 2.95     & 3.15    \\ \bottomrule
\end{tabular}%
}
\end{table}

Bus routes service hours contribute significant power in predicting ridesourcing demand at both contexts, which is consistent with \cite{yan2020using}. Besides bus routes, rail station density also plays a significant role. Notably, it has the largest importance (7.7\%) on airport trips, while only 2.44\% for downtown trips and 1.45\% for neighborhood trips. A possible explanation is that people may prefer taking rail transportation for airport trips which has a relatively longer travel distance.

Regarding the land use and accessibility variables, employment density displays the most important across three trip types, followed by intersection density and retail density. Employment density is the most important attribute among built environment. This finding is consistent with other studies that employment is highly related to travel demand \citep{chen2019discovering, chen2021nonlinear, sabouri2020exploring}. Employment density achieves the greatest importance for downtown trips, followed by neighborhood and airport. This finding highlights the role of ridesourcing in attracting riders for business trip purposes. Population density does not show a significant influence in ridesourcing demand for downtown and neighborhood trips. This finding is in line with prior research that population density has a relatively lower influence on ridesourcing demand than employment density \citep{marquet2020spatial, yan2020using}. Intersection density and retail employment density also enjoy varying predictive importance among three types of trips. They both rank the highest in neighborhood context and followed by downtown context.

Commute variables rank quite high in both contexts. Since ridesourcing services are gradually integrated into daily commuting, thus the size of commuting flow has strong contributions in influencing ridesourcing demand. The commuter-related variables have the most importance in forecasting downtown ridesourcing demand. Besides, the importance value of \textit{Commuters\_HW} ranks higher than \textit{Commuters\_WH}, suggesting home-to-work commutes are more important in influencing ridesourcing demand than work-to-home commutes. 

Moreover, we find that travel impedance variables have a substantial importance value in both contexts. This underscores that travel cost is a dominating factor in forecasting ridesourcing demand. Also, travel impedance variables have strong heterogeneous effects. For example, \textit{Fare\_sd} has larger importance in neighborhood context than others. This suggests that travel cost variation has relatively more influence on neighborhood ridesourcing trips. \textit{Fare\_median} plays important roles in shaping both downtown and neighborhood trips but is less important for airport trips, indicating that travelers who take ridesourcing for airport rides are less sensitive to the trip cost.

We also calculated the relative variable importance aggregated and averaged by variable category, as shown in Table \ref{tab:category_importance}. Socioeconomic and demographic variables are important in predicting travel demand \citep{ewing2010travel, yan2020using}. Our study confirms this finding. The results show that this variable category collectively accounts for a significant proportion of importance (around 40\%) across three contexts. The sum of variable importance for built environment variables displays an obvious heterogeneity across contexts. Specifically, built environment has the most contributing power (50.35\%) in predicting the demand for airport rides, followed by 47.28\% for downtown rides and the least (40.64\%) for neighborhood rides. Despite the heterogeneity, the results collectively show that built environment is more important than socioeconomic and demographic variables, which is consistent with previous travel demand modeling studies \citep{chen2021nonlinear, tu2021exploring, shao2020threshold}. This finding further highlights the critical role of built environment on shaping travel behavior \citep{ewing2010travel}. By contrast to the joint contribution, socioeconomic and demographic variables are averagely more important than built environment over all contexts. Which of these two variable categories has a more important role on travel behavior has been a long-standing debate, and has been analyzed for different outcomes including travel demand modeling \citep{chen2021nonlinear, yan2020using, ding2019does}, commuting mode choice \citep{ding2018synergistic}, driving distance \citep{ding2018applying} and car ownership \citep{zhang2020nonlinear}. Our study contributes a piece of evidence of their predictive importance in ridesourcing demand modeling. The heterogeneous effects of the travel impedance variables are evident. They account for the most importance (18.97\%) in shaping neighborhood ridesourcing demand, followed by downtown (13.71\%) and then airport (8.73\%). And averagely, they have the greatest predictive power to the ridesourcing demand, which echoes the findings in \citep{yan2020using, xu2021identifying, tu2021exploring}. This further underscores the importance of including travel impedance variables in travel demand modeling.

\subsection{Nonlinear Relationships}

\begin{figure}[!htbp]
    \centering
    \includegraphics[width=\textwidth]{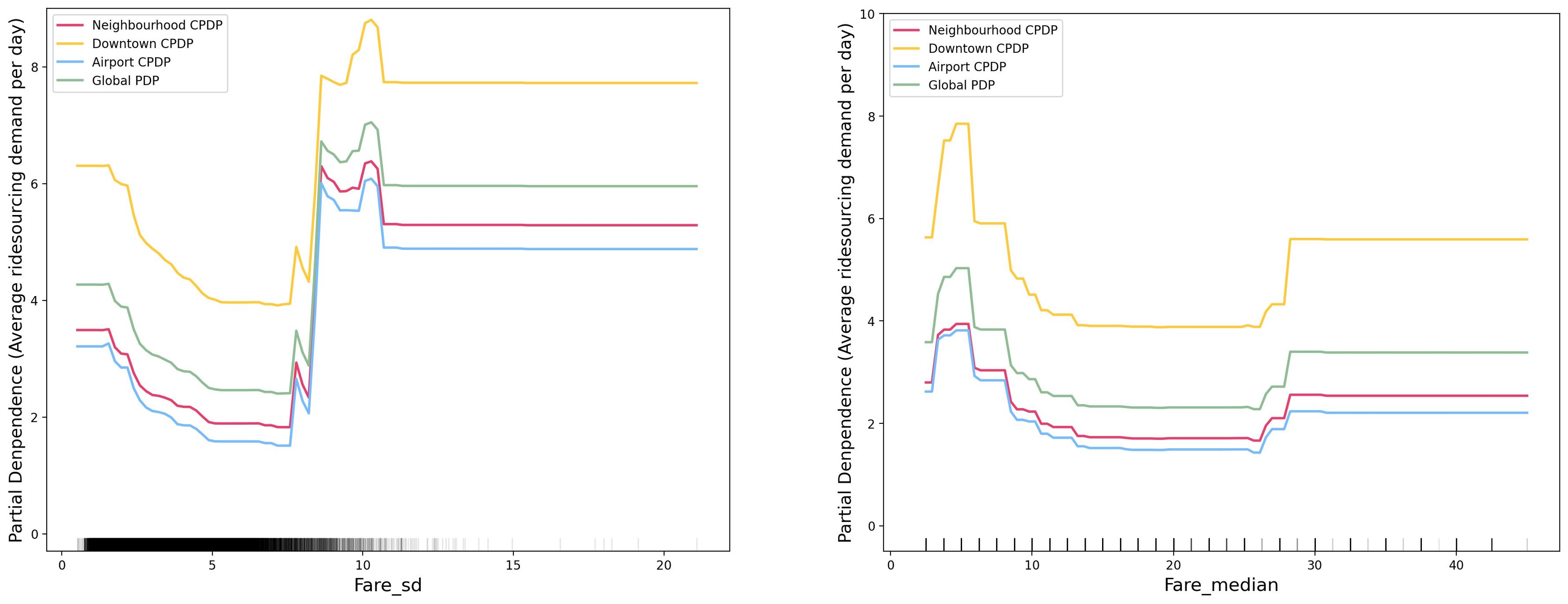}
    \caption{Trip cost}
    \label{fig:cpdp_fare}
\end{figure}

Although the variable importance has proved that variables have heterogeneous importance under different contexts, we can hardly decide their associations with the ridesourcing demand. To address this issue, this study adopts CPDP to examine the nonlinear relationships across different contexts. The results of CPDPs for several key variables are presented in Fig. \ref{fig:cpdp_fare}. We also present the global PDPs for comparison. The rug marks at the bottom of each plot show the distribution of the studied variable; and note that for ranges where the marks are sparse, the results may be less reliable and thus require careful consideration.

For \textit{Fare\_sd}, the PDP and CPDP both display a strong nonlinear relationship (i.e., U-shape) with the ridesourcing demand. Specifically, when \textit{Fare\_sd}, increases from around 1 to 5, all curves experience a gradual decrease. This is intuitive since the rising fare uncertainty (caused by travel distance or time uncertainty) may lower people’s willingness to use ridesourcing services. The curve in Downtown Context is higher than in the other two contexts, demonstrating that downtown travel contributes the most to the ridesourcing ridership. When \textit{Fare\_sd} grows from 5 to 8, the ridesourcing demand remains flat, indicating that when the travel fare variation falls into a range, travelers will be less sensitive to the trip cost. One possible explanation here is that the increasing travel demand may counter the decline of ridesourcing demand caused by fare uncertainty. To be specific, fare variation may be mainly caused by traffic congestion, which usually happens in areas with higher travel demand. When \textit{Fare\_sd} exceeds 8, the ridesourcing demand shows a surge, suggesting that the increased fare variation even brings more ridesourcing usage. One important fact related to this finding is that more than 75\% OD pairs with \textit{Fare\_sd} exceeding eight are in airport context. We believe that these airport rides are the major mediating factor underlying this relationship. Previous studies found that the taxi adoption rate for airport rides in Chicago is relatively low and transit coverage for airport trips is poor \citep{soria2020k}; and travelers are more likely to use ridesourcing for airport trips \citep{alemi2019drives}. \ref{tab:category_importance} also shows travel impedance has the least predictive importance in airport context. Consequently, even if \textit{Fare\_sd} keeps growing, the ridesourcing demand still shows an increasing trend. Finally, the \textit{Fare\_sd} displays a saturated effect on ridesourcing demand when it surpasses 10, probably due to insufficient data points. Overall, ridesourcing demand decreases or increases more drastically in Downtown Context. This suggests that ridesourcing users in downtown area are more sensitive to travel cost uncertainty. The \textit{Fare\_median} also displays a similar pattern as \textit{Fare\_sd}. And we can more easily observe that downtown ridesourcing demand has the highest sensitivity to travel cost changes.

\begin{figure}[!htbp]
    \centering
    \includegraphics[width=\textwidth]{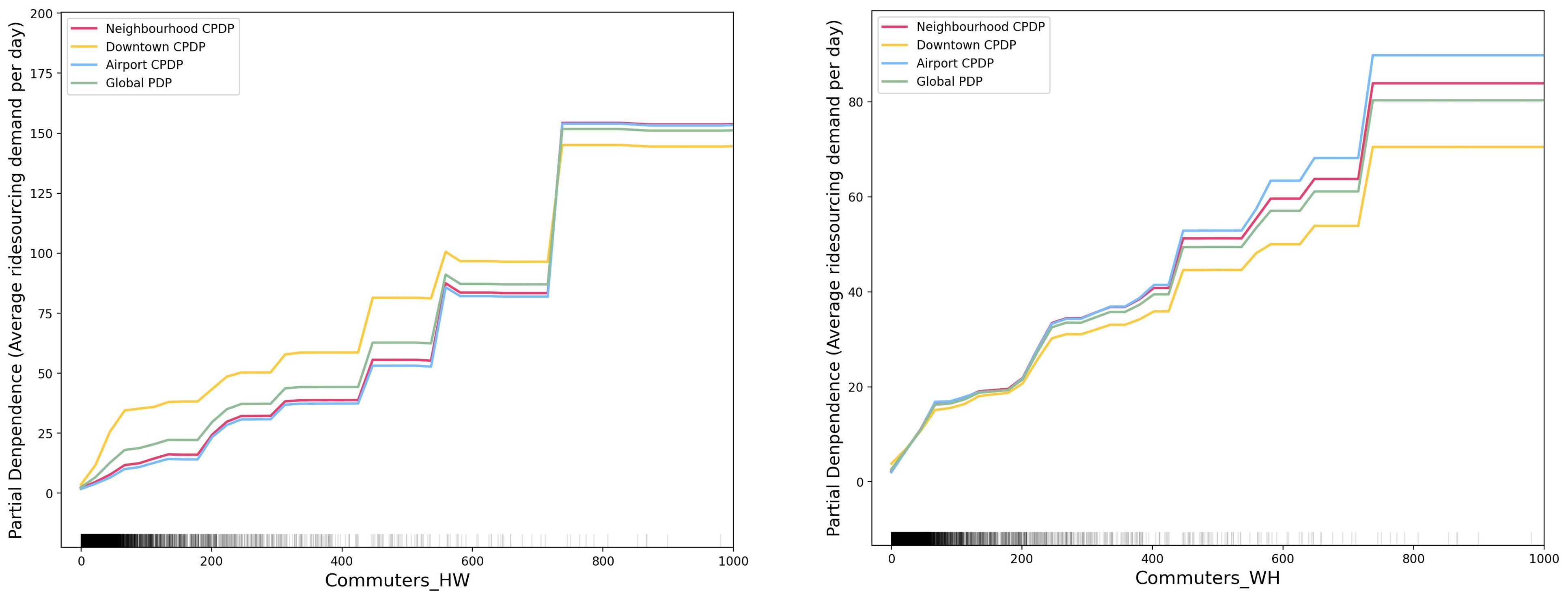}
    \caption{The number of commuters}
    \label{fig:cpdp_commuters}
\end{figure}

Fig. \ref{fig:cpdp_commuters} shows that the number of \textit{Commuters\_HW} and \textit{Commuters\_WH} present heterogeneous effects on ridesourcing demand across three trip types. The presented curves demonstrate that the home-to-work commutes have a stronger positive effect on ridesourcing demand. Increasing the number of home-to-work commuters from the range with enough data points (i.e., 0 to 400) could generate around 50 ridesourcing trips in downtown and 25 trips to airport and neighborhoods. However, increasing the same amount of work-to-home commuters can only yield around 30 trips for downtown context and around 40 for the others. One plausible reason for such difference is that home-to-work commute and work-to-home commute tend to happen during morning and afternoon peaks, respectively; and people may have different sensitivity to time flexibility during these two periods \citep{oakil2016rush}. For example, people may have various activities such as shopping or recreations when going home due to greater time flexibility, thus using ridesourcing may be less prioritized; however, there is limited time flexibility allowed when going to work during morning peak, hence people may rely more on ridesourcing because of its time-efficient characteristic compared with public transit \citep{soria2020k}. Ridesourcing demand experiences greater increases in downtown context with the increase of \textit{Commuters\_HW} compared with \textit{Commuters\_WH}, indicating that the ridesourcing usage is especially more sensitive to the commuters who work in downtown areas during home-to-work commuting. Note that when the number of commuters reaches beyond 600, the prediction would be relatively unreliable due to limited data. Consistent with the results of variable importance, the largest influence of socioeconomic and demographic variables comes from commutes variables.

\begin{figure}[!ht]
    \centering
    \includegraphics[width=\textwidth]{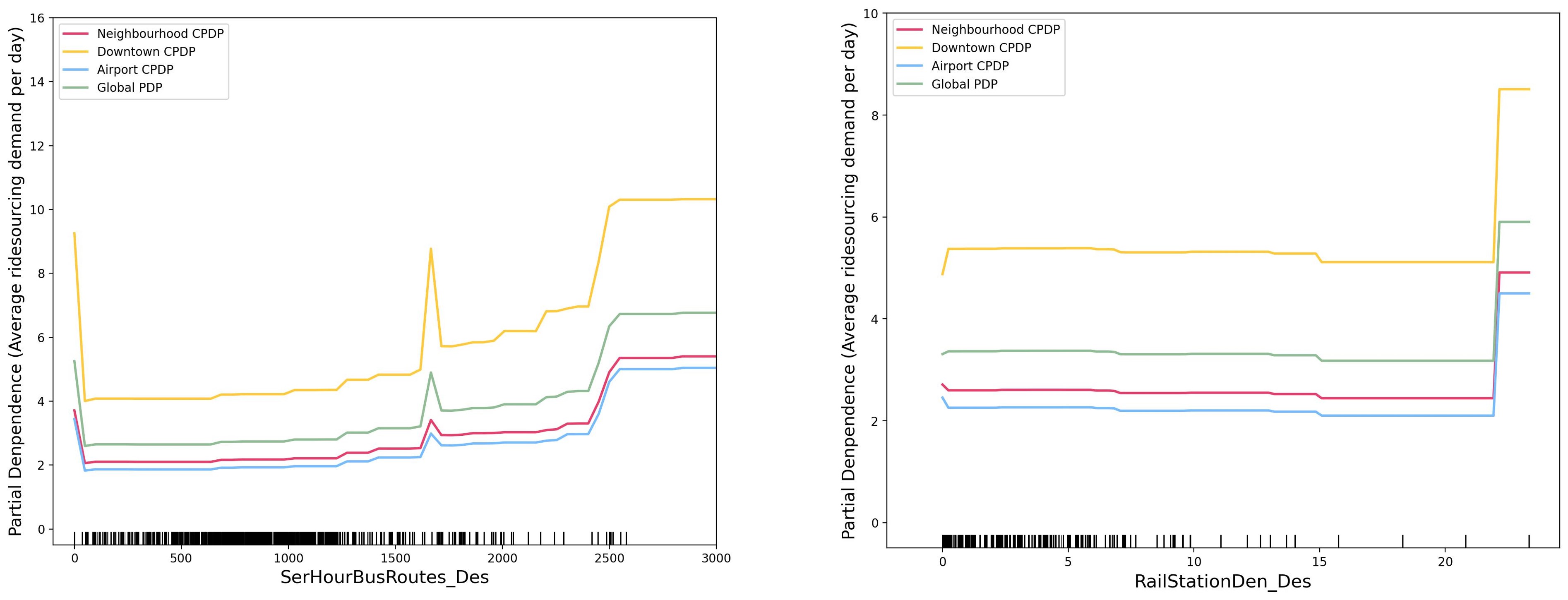}
    \caption{Key transit supply variables}
    \label{fig:cpdp_transit}
\end{figure}

Fig. \ref{fig:cpdp_transit} reveals the nonlinear effects of aggregated bus routes service hour on ridesourcing demand across contexts. The ridesourcing demand experiences a sharp decrease in all contexts when \textit{SerHourBusRoutes} increases from 0 to around 100. This finding is more evident in downtown context, as the ridesourcing demand drops more than 2 times. This finding further confirms that ridesourcing plays a complementary role in areas where public transit services are less frequent \citep{jin2018ridesourcing}. When there are no transit services, ridesourcing may become the preferred option for traveling. Ridesourcing demand slowly rises when \textit{SerHourBusRoutes} increases from 100 to around 1500. This is probably because ridesourcing usually serves as feeder service for public transit \citep{zgheib2020modeling}. When \textit{SerHourBusRoutes} exceeds around 2500, the ridesourcing demand almost saturates. Beyond 1500, the reliability of the modeling results may gradually be compromised by fewer data points. Even if all curves share the same effective range and thresholds, the change in ridesourcing demand is inconsistent. Among which, \textit{SerHourBusRoutes} is the most sensitive to downtown ridesourcing riders, followed by neighborhood and airport. \textit{RailStationDen} shows a relatively less effect on ridesourcing demand across all types of trips, showing that increasing rail stop density will barely attract ridesourcing adoptions. It has a negative effect on ridesourcing demand when the density is from 0 to 10. This finding is consistent with \citep{ghaffar2020modeling} that increasing rail transit services tends to generate fewer ridesourcing trips.

\begin{figure}[!ht]
    \centering
    \includegraphics[width=\textwidth]{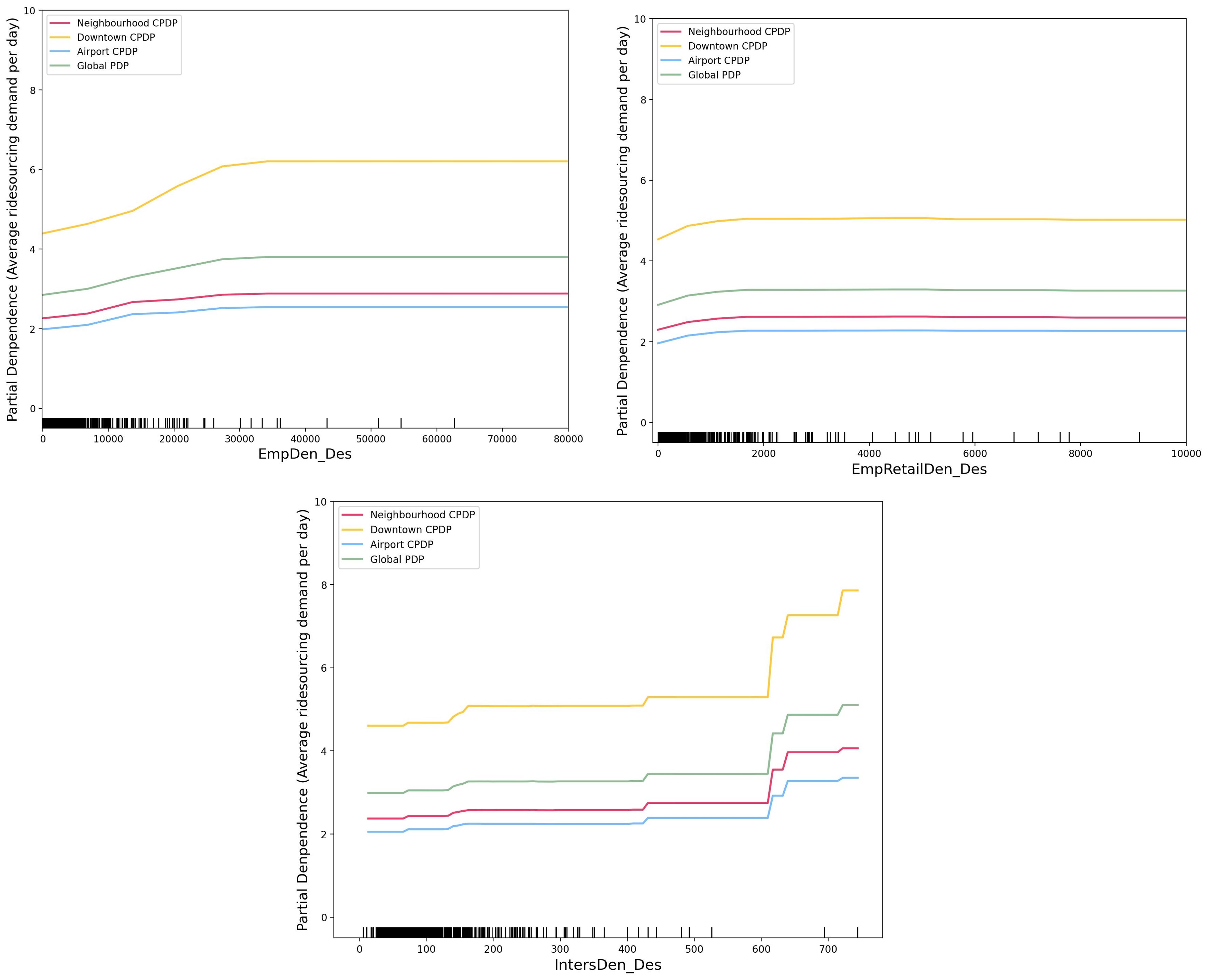}
    \caption{Key land use and accessibility variables}
    \label{fig:cpdp_emp}
\end{figure}

Fig. \ref{fig:cpdp_emp} presents the nonlinear relationships between three key built environment variables and ridesourcing demand. Compared with transit supply variables, the effects of employment is limited. The largest effect, which happens in downtown context, is only around 2. Employment density has a positive influence on ridesourcing demand, which echoes the findings from prior studies \citep{yan2020using, ghaffar2020modeling}. Ridesourcing demand increases modestly in airport and neighborhood context when employment density is within the effective range, i.e., 0 to around 28,000 $jobs/mi^2$. After this threshold, the effects are saturated. Employment density has a more intense effect in shaping downtown ridesourcing trips. The curve is relatively sharper when employment density rises from 0 to around 28,000 $jobs/mi^2$. The effect is saturated when employment density goes beyond 28,000 $jobs/mi^2$. After that, the estimation becomes weaker due to insufficient data. The effects of retail employment density are almost negligible (i.e., the largest effect is only 1) in all contexts; and all curves share the same effective range. The effects are almost saturated at a threshold of around 2000 $retail jobs/mi^2$. Moreover, the change in ridesourcing demand shows that the influence of retail employment density is stronger in downtown context than in the others. The intersection density is positively related to ridesourcing demand, which echoes the findings from previous studies of travel demand modeling  \citep{yu2019exploring, chen2021nonlinear, ding2019does}. The effects of intersection density are marginal compared with employment density. The curves remain almost flat before intersection density reaches 140. After this threshold, ridesourcing demand experiences a slight increase and then remains silent. Before 300 $intersections/mi^2$, the estimates are supported by enough data. Even if all curves have a rise in 600 $intersections/mi^2$, the reliability of the prediction may be compromised with fewer data. Also, the effects of intersection density vary across different types of trips. Within the range of 0 to 300 $intersections/mi^2$, the effects are largest in downtown context, followed by neighborhood and airport context.

\section{Conclusion}
\label{discussion and conclusion}

To summarize, this study develops an explainable machine learning framework for investigating which factors shape ridesourcing demand and examining their nonlinear and spatially heterogeneous effects. Specifically, we examine how variable importance (for predicting ridesourcing trips) and identified nonlinear effects differ across the downtown, airport and neighborhood contexts. Our work thus extends previous ridesourcing studies which have mostly assumed linear and spatially homogeneous relationships. 



The results showed that built environment variables collectively contribute more than socioeconomic and demographic variables to predict ridesourcing demand across all three spatial contexts (airport, downtown, and neighborhood). However, individual socioeconomic and demographic variables have higher importance values on average. This finding highlights the importance of both built environment and sociodemographic variables in shaping ridesourcing demand. Among transit supply variables, bus routes service hour contributes the most in predicting downtown trips whereas rail station density is the most important variable for predicting airport trips. These findings suggest that there is spatial heterogeneity regarding the influence of transit supply on ridesourcing demand. Among land use and accessibility variables, employment density has the most predictive power for downtown and neighborhood trips, while population density is the most important variable in predicting airport trips. Moreover, we find spatial heterogeneity in travel impedance variables (especially travel cost) with regard to variable importance. Specifically, it has the most contributing power in predicting neighborhood trips, followed by downtown; but it is less sensitive to airport rides. 



The \textit{Conditional Partial Dependence Plots} (CPDPs) further reveal that the nonlinear relationships between built-environment variables, including employment density, retail density, intersection density, bus route service hours, and rail station density and ridesourcing demand. Regarding spatial heterogeneity, we find that ridesourcing demand is more responsive to transit supply changes in the downtown context than in the airport or neighborhood context. We also find that even if the nonlinear effects (i.e., effective range and thresholds) of the built environment variables on ridesourcing demand display similar patterns across various spatial contexts, the magnitudes of change in ridesourcing demand are inconsistent. The magnitude is usually the largest for downtown trips, followed by neighborhood trips and airport trips. Among all independent variables, changes in the number of commuters lead to the largest magnitude of change in ridesourcing demand. In addition, CPDP curves suggest that travel impedance variables have nonlinear associations with ridesourcing demand and that ridesourcing demand is more sensitive to travel cost changes in downtown than in the airport or neighborhood context. These findings offer strong empirical evidence that the associations of ridesourcing demand and its determinants are both nonlinear and spatially heterogeneous. Uncovering these complex relationships enables transportation planners and policymakers to develop more context-specific, targeted policies and strategies to manage ridesourcing services.

A notable limitation of this study is that we have not included temporal and weather variables in the ridesourcing demand model. Previous studies have found that these variables could significantly impact ridesourcing demand \citep{ghaffar2020modeling}. Future research may extend our work, which focuses on examining spatial patterns, by further considering temporal and weather variables to uncover the spatiotemporal dynamics in ridesourcing trip generation. Moreover, future work may extend our explainable machine learning analytical framework to examine ridesourcing services in other cities or to examine other travel modes. Additional empirical analysis and further validation of the proposed analytical approach can enhance the transferability of the study findings.


\section*{Authorship Contribution Statement}
\textbf{Zhang}: Conceptualization, Data Curation, Methodology, Software, Formal Analysis, and Draft Preparation. \textbf{Yan}: Conceptualization, Formal Analysis, Draft Preparation, Supervision and Grant Acquisition. \textbf{Zhou}: Conceptualization, Methodology, Formal Analysis, Draft Preparation. \textbf{Xu}: Data Curation, Visualization, Formal Analysis. \textbf{Zhao}: Conceptualization, Methodology, Draft Preparation, Supervision and Grant Acquisition.

\section*{Acknowledgment}

This research was partially supported by the U.S. Department of Transportation through the Southeastern Transportation Research, Innovation, Development and Education (STRIDE) Region 4 University Transportation Center (Grant No. 69A3551747104). 
\newpage

\FloatBarrier

\section*{Appendix}
\label{Appendix}

\begin{table}[!ht]
\centering
\footnotesize
\caption{Descriptive Statistics}
\label{tab:Descriptive Statistics}
\resizebox{\textwidth}{!}{%
\begin{tabular}{lllllll}
\hline
                                       & \multicolumn{2}{l}{Airport} & \multicolumn{2}{l}{Downtown} & \multicolumn{2}{l}{Neighborhood} \\ \cline{2-7} 
                                       & Mean         & Std          & Mean         & Std           & Mean            & Std            \\ \hline
\textbf{Target variable}               &              &              &              &               &                 &                \\
Average number of trips per day        & 8.11         & 24.65        & 6.40         & 21.90         & 1.61            & 2.23           \\
\textbf{Travel impedance}                       &              &              &              &               &                 &                \\
Fare\_median                           & 22.16        & 7.53         & 11.50        & 3.68          & 7.59            & 3.01           \\
Fare\_sd                               & 6.33         & 2.09         & 3.80         & 1.23          & 2.44            & 0.96           \\
\textbf{Socioeconomic and demographic} &              &              &              &               &                 &                \\
Pctsinfam\_Ori                         & 0.14         & 0.22         & 0.13         & 0.16          & 0.24            & 0.20           \\
Pctsinfam\_Des                         & 0.12         & 0.20         & 0.12         & 0.15          & 0.24            & 0.20           \\
Pcthisp\_Ori                           & 0.13         & 0.23         & 0.15         & 0.20          & 0.23            & 0.25           \\
Pcthisp\_Des                           & 0.11         & 0.21         & 0.14         & 0.19          & 0.22            & 0.25           \\
Pctasian\_Ori                          & 0.04         & 0.08         & 0.11         & 0.10          & 0.07            & 0.09           \\
Pctasian\_Des                          & 0.04         & 0.08         & 0.12         & 0.10          & 0.07            & 0.09           \\
Pctlowinc\_Ori                         & 0.14         & 0.17         & 0.21         & 0.13          & 0.26            & 0.16           \\
Pctlowinc\_Des                         & 0.12         & 0.16         & 0.20         & 0.13          & 0.26            & 0.16           \\
CrimeDen\_Ori                          & 83.71        & 139.91       & 242.69       & 263.27        & 149.35          & 138.19         \\
CrimeDen\_Des                          & 68.81        & 125.36       & 243.46       & 263.51        & 145.18          & 134.10         \\
Commuters\_HW                          & 5.98         & 11.70        & 17.00        & 60.31         & 2.68            & 7.32           \\
Commuters\_WH                          & 5.53         & 11.60        & 16.59        & 60.39         & 2.59            & 7.28           \\
\textbf{Built environment}             &              &              &              &               &                 &                \\
\textit{Land use and accessibility}    &              &              &              &               &                 &                \\
Popden\_Ori                            & 11464.10     & 16170.07     & 28578.29     & 19884.29      & 20449.43        & 12706.20       \\
Popden\_Des                            & 10447.98     & 16286.36     & 28823.54     & 19618.55      & 20276.88        & 13061.63       \\
IntersDen\_Ori                         & 63.94        & 76.88        & 188.59       & 157.04        & 110.05          & 73.80          \\
IntersDen\_Des                         & 59.38        & 76.50        & 192.26       & 159.07        & 110.73          & 74.85          \\
EmpDen\_Ori                            & 7355.76      & 26934.16     & 67875.80     & 132529.45     & 5835.92         & 8052.61        \\
EmpDen\_Des                            & 7587.41      & 26900.72     & 71667.34     & 134899.91     & 6018.72         & 8245.48        \\
EmpRetailDen\_Ori                      & 517.42       & 2531.34      & 4444.46      & 9791.31       & 715.91          & 1244.74        \\
EmpRetailDen\_Des                      & 503.61       & 2530.12      & 4694.73      & 10044.96      & 711.25          & 1237.12        \\
\textit{Transit supply}                &              &              &              &               &                 &                \\
SerHourBusRoutes\_Ori                  & 760.58       & 613.90       & 1943.09      & 1261.49       & 1042.66         & 452.12         \\
SerHourBusRoutes\_Des                  & 747.41       & 624.20       & 1990.23      & 1267.97       & 1052.23         & 457.71         \\
PctRailBuf\_Ori                        & 0.12         & 0.22         & 0.36         & 0.35          & 0.20            & 0.29           \\
PctRailBuf\_Des                        & 0.11         & 0.21         & 0.37         & 0.35          & 0.21            & 0.29           \\
BusStopDen\_Ori                        & 37.44        & 33.79        & 80.73        & 47.46         & 60.43           & 26.58          \\
BusStopDen\_Des                        & 34.59        & 33.42        & 81.58        & 47.70         & 60.35           & 26.83          \\
RailStationDen\_Ori                    & 0.75         & 2.04         & 2.94         & 5.00          & 1.23            & 2.70           \\
RailStationDen\_Des                    & 0.74         & 2.05         & 3.02         & 5.06          & 1.25            & 2.72           \\ \hline
\end{tabular}%
}
\end{table}

\FloatBarrier
\newpage





\bibliographystyle{elsarticle-harv}
\biboptions{semicolon,round,sort,authoryear}
\bibliography{sample.bib}







\end{document}